\documentclass[a4paper,12pt]{article}
\usepackage{nldb}
\usepackage{epsfig}

\title{STRUCTURED LANGUAGE MODELING FOR SPEECH RECOGNITION
\footnote{This work was funded by the NSF IRI-19618874 grant STIMULATE}}
\author{Ciprian Chelba and Frederick Jelinek}
\summary{A new language model for speech recognition is presented. 
The model develops hidden hierarchical syntactic-like structure
incrementally and uses it to extract meaningful information from the
word history, thus complementing the locality of currently used trigram models.
The structured language model (SLM) and its performance in a two-pass
speech recognizer --- lattice decoding ---  are
presented. Experiments on the WSJ corpus show an improvement 
in both perplexity (PPL) and word error rate (WER) over conventional trigram
models.}

\begin{document}
\maketitle
\setlength{\baselineskip}{0.6cm}
\section{Structured Language Model}

\begin{figure}[h]
  \begin{center}
    \epsfig{file=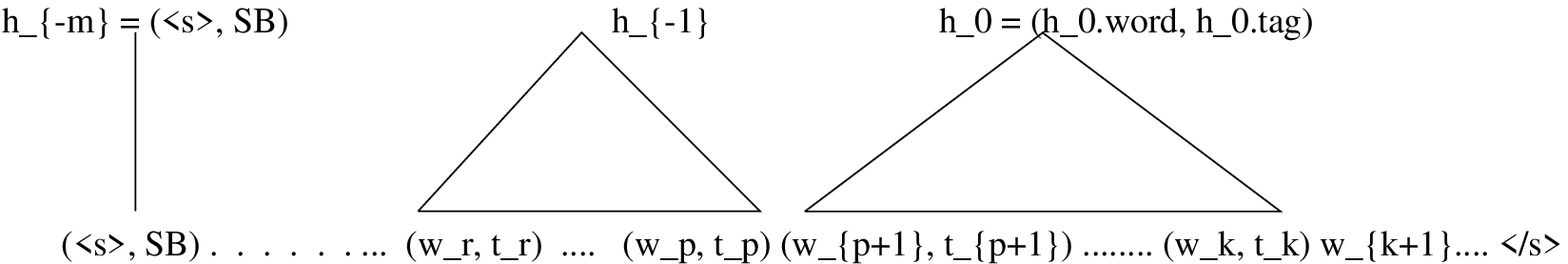,height=1.5cm,width=7cm}
  \end{center}\vspace{-0.9cm}
  \mycaption{A word-parse k-prefix} \label{fig:w_parse}
\end{figure}

\noindent An extensive presentation of the SLM can be found
in~\cite{chelba98}. The model assigns a probability $P(W,T)$ to every
sentence $W$ and its every possible binary parse $T$. The
terminals of $T$ are the words of $W$ with POStags, and the nodes of $T$ are
annotated with phrase headwords and non-terminal labels.\\
 Let $W$ be a sentence of length $n$ words to which we have prepended
\verb+<s>+ and appended \verb+</s>+ so that $w_0 = $\verb+<s>+ and
$w_{n+1} = $\verb+</s>+.
Let $W_k$ be the word k-prefix $w_0 \ldots w_k$ of the sentence and 
\mbox{$W_k T_k$} the \emph{word-parse k-prefix}. Figure~\ref{fig:w_parse} shows a
word-parse k-prefix; \verb|h_0 .. h_{-m}| are the \emph{exposed
 heads}, each head being a pair(headword, non-terminal label), or
(word,  POStag) in the case of a root-only tree. 

\subsection{Probabilistic Model} \label{section:prob_model}

\begin{figure}
  \begin{center} 
    \epsfig{file=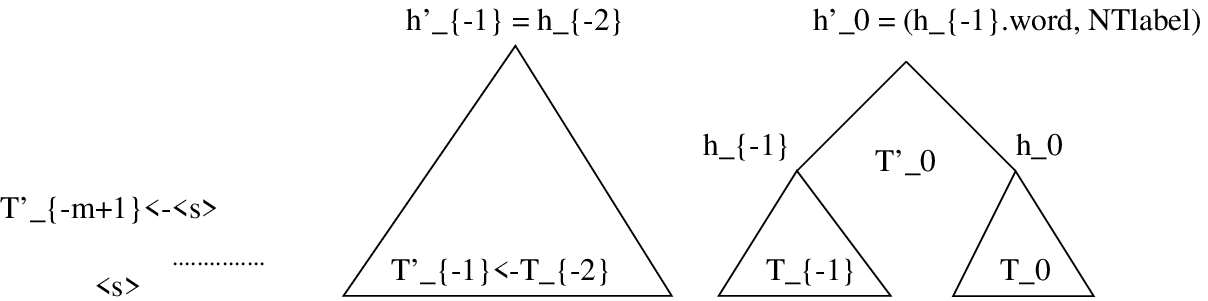,height=1.7cm,width=7cm}
  \end{center}\vspace{-0.9cm}
  \mycaption{Result of adjoin-left under NTlabel} \label{fig:after_a_l}
\end{figure}

\begin{figure}
  \begin{center} 
    \epsfig{file=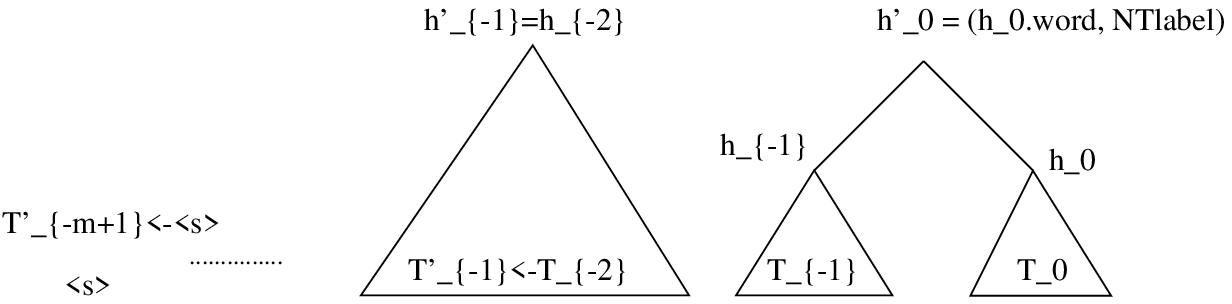,height=1.7cm,width=7cm}
  \end{center}\vspace{-0.9cm}
  \mycaption{Result of adjoin-right under NTlabel} \label{fig:after_a_r}
\end{figure}

\noindent The probability $P(W,T)$ of a word sequence $W$ and a complete parse
$T$ can be broken into:\vspace{-0.3cm}
\begin{eqnarray*}
P(W,T) & = & \prod_{k=1}^{n+1}[ P(w_k/W_{k-1}T_{k-1}) \cdot P(t_k/W_{k-1}T_{k-1},w_k) \cdot
\prod_{i=1}^{N_k}P(p_i^k/W_{k-1}T_{k-1},w_k,t_k,p_1^k\ldots p_{i-1}^k)]
\end{eqnarray*}
where: \\
$\bullet$ $W_{k-1} T_{k-1}$ is the word-parse $(k-1)$-prefix\\
$\bullet$ $w_k$ is the word predicted by WORD-PREDICTOR\\
$\bullet$ $t_k$ is the tag assigned to $w_k$ by the TAGGER\\
$\bullet$ $N_k - 1$ is the number of operations the PARSER executes at 
sentence position $k$ before passing control to the  WORD-PREDICTOR
(the $N_k$-th operation at position k is the \verb+null+ transition);
$N_k$ is a function of $T$\\
$\bullet$ $p_i^k$ denotes the i-th PARSER operation carried out at
position k in the word string; the operations performed by the
PARSER are illustrated in
Figures~\ref{fig:after_a_l}-\ref{fig:after_a_r} and they ensure that
all possible binary branching parses with all possible headword and
non-terminal label assignments for the $w_1 \ldots w_k$ word
sequence can be generated.
 Our model is based on three probabilities, estimated using deleted
interpolation (see~\cite{jelinek80}), parameterized as follows:
\vspace{-0.3cm}
\begin{eqnarray}
  P(w_k/W_{k-1} T_{k-1}) & = & P(w_k/h_0, h_{-1})\label{eq:1}\\
  P(t_k/w_k,W_{k-1} T_{k-1}) & = & P(t_k/w_k, h_0.tag, h_{-1}.tag)\label{eq:2}\\
  P(p_i^k/W_{k}T_{k}) & = & P(p_i^k/h_0, h_{-1})\label{eq:3}
\end{eqnarray}
 It is worth noting that if the binary branching structure
developed by the parser were always right-branching and we mapped the
POStag and non-terminal label vocabularies to a single type then our
model would be equivalent to a trigram language model.\\
 Since the number of parses  for a given word prefix $W_{k}$ grows
exponentially with $k$, $|\{T_{k}\}| \sim O(2^k)$, the state space of
our model is huge even for relatively short sentences so we had to use
a search strategy that prunes it. Our choice was a synchronous
multi-stack search algorithm which is very similar to a beam search. \\
 The probability assignment for the word at position $k+1$ in the input sentence is made using:
\vspace{-0.3cm}
\begin{eqnarray}
  P(w_{k+1}/W_{k}) & = & \sum_{T_{k} \in S_{k}}P(w_{k+1}/W_{k}T_{k})\cdot
[\ P(W_{k}T_{k})/\sum_{T_{k} \in S_{k}}P(W_{k}T_{k})\ ]
\end{eqnarray}
which ensures a proper probability over strings $W^*$, where $S_{k}$ is
the set of all parses present in our stacks at the current stage $k$.
An N-best EM variant is employed to reestimate the model parameters
such that the PPL on training data is decreased --- the likelihood of
the training data under our model is increased. The reduction
in PPL is shown experimentally to carry over to the test data.

\section{$A^*$ Decoder for Lattices}

\noindent The \emph{speech recognition lattice} is an intermediate format in which the hypotheses
produced by the first pass recognizer are stored. For each utterance
we save a \emph{directed acyclic graph} in which the \emph{nodes} are
a subset of the language model states in the composite hidden Markov
model and the arcs --- \emph{links} --- are labeled with
words. Typically, the first pass acoustic/language model scores
associated with each link in the lattice are saved and the nodes
contain time alignment information.

\noindent There are a couple of reasons that make $A^*$~\cite{astar} appealing for
lattice decoding using the SLM:\\
$\bullet$ the algorithm operates with whole prefixes, making
  it ideal for incorporating language models whose memory is the
  entire sentence prefix;\\
$\bullet$ a reasonably good lookahead function and an efficient way to
  calculate it using dynamic programming techniques are both readily
  available using the n-gram language model.

\subsection{$A^*$ Algorithm}
\noindent Let a set of hypotheses $L=\{h:x_1,\ldots, x_n\},\ x_i \in \mathcal{W}^*\ \forall\ i$
be organized as a prefix tree. We wish to obtain the maximum scoring
hypothesis under the scoring function $f:\mathcal{W}^* \rightarrow \Re$: 
$h^*=\arg\max_{h \in L}f(h)$ without scoring all the hypotheses in
$L$, if possible with a minimal computational effort.
The $A^*$ algorithm operates with prefixes and suffixes of hypotheses ---
paths --- in the set $L$; we will denote prefixes --- anchored at the root of the tree
--- with $x$ and suffixes --- anchored at a leaf --- with $y$. A
complete hypothesis $h$ can be regarded as the concatenation of a $x$
prefix and a $y$ suffix: $h = x.y$. 

\begin{figure}[htbp]
  \begin{center}
    \epsfig{file=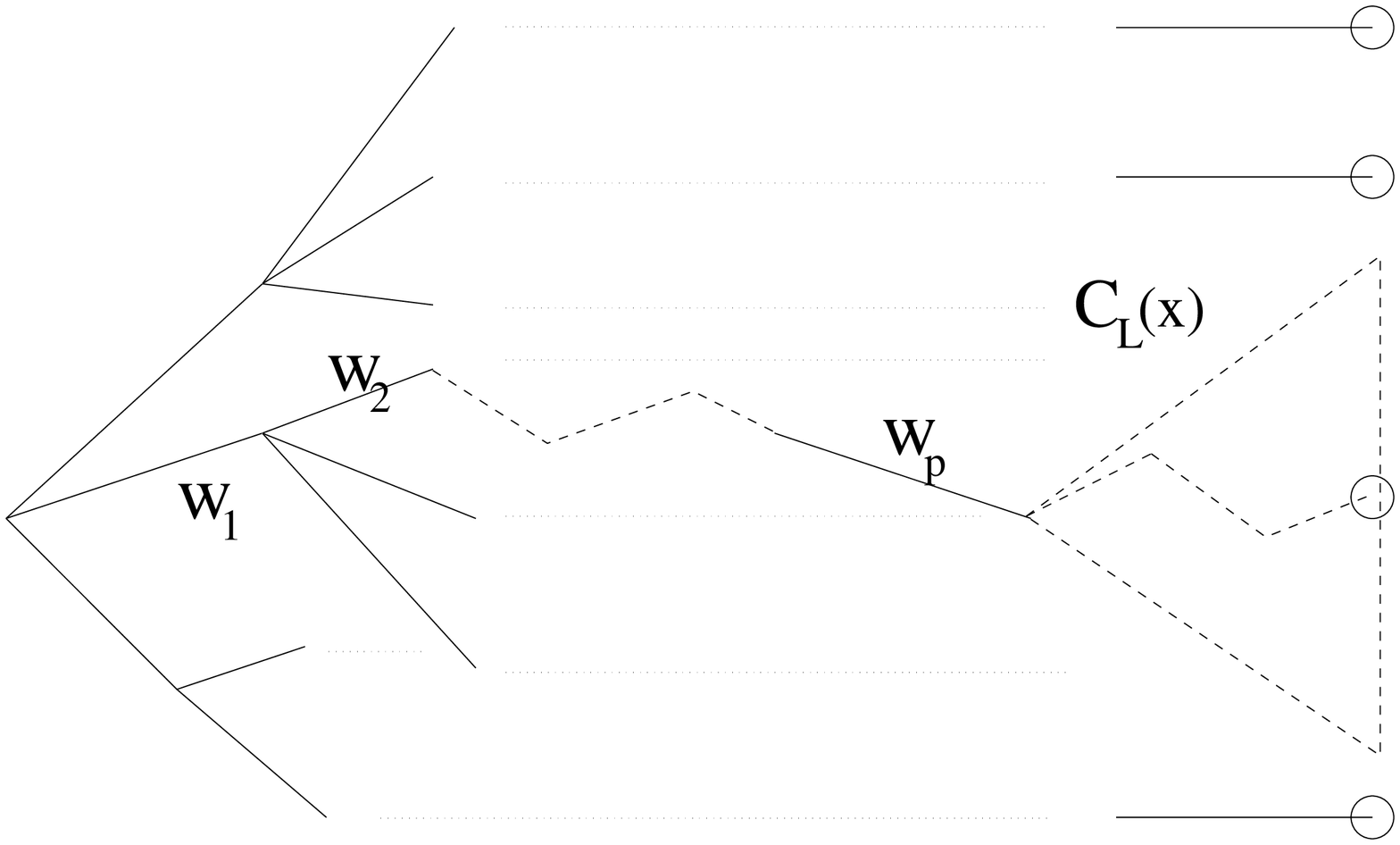, height=3.5cm,width=8cm}\vspace{-0.5cm}
    \mycaption{Prefix Tree Organization of a Set of Hypotheses L}
    \label{fig:prefix_tree}
  \end{center}
\end{figure}

\noindent To be able to pursue the most promising path, the algorithm needs to
evaluate all the possible suffixes that are allowed in $L$ for a given prefix
$x=w_1,\ldots,w_p$ --- see Figure~\ref{fig:prefix_tree}.  Let $C_L(x)$
be the set of suffixes allowed by the tree for a prefix $x$ and assume
we have an overestimate for the $f(x.y)$ score of any \emph{complete}
hypothesis $x.y$: $g(x.y) \doteq f(x) + h(y|x) \geq f(x.y)$.
Imposing that $h(y|x) = 0$ for empty $y$, we have
$g(x) = f(x), \forall\ complete\ x \in L$ that is, the overestimate 
becomes exact for complete hypotheses $h \in L$.
Let the \emph{$A^*$ ranking function} $g_L(x)$ be:
\vspace{-0.3cm}
\begin{eqnarray}
  g_L(x) & \doteq & \max_{y \in C_L(x)} g(x.y) = f(x) + h_L(x),\ where
  \label{eq:g_overestimate}\\
  h_L(x) & \doteq & \max_{y \in C_L(x)} h(y|x)
\end{eqnarray}
$g_L(x)$ is an overestimate for the $f(\cdot)$ score of any complete
hypothesis that has the prefix $x$; the overestimate becomes exact for
complete hypotheses.
The $A^*$ algorithm uses a potentially infinite stack in which prefixes $x$ are
ordered in decreasing order of the $A^*$ ranking function
$g_L(x)$;
at each extension step the top-most prefix $x=w_1,\ldots,w_p$ is popped
from the stack, expanded with all possible one-symbol continuations of
$x$ in $L$ and then all the resulting expanded prefixes --- among
which there may be complete hypotheses as well --- are inserted back
into the stack. The stopping condition is: whenever the popped
hypothesis is a complete one, retain it as the overall best
hypothesis $h^*$.

\subsection{$A^*$ Lattice Rescoring} \label{section:lat_rescoring}

\noindent A speech recognition lattice can be conceptually organized as a
prefix tree of paths. When rescoring the lattice using a different language model than the
one that was used in the first pass, we seek to find the complete path 
$p=l_0 \ldots l_n$ maximizing:
\vspace{-0.3cm}
\begin{equation}
  \label{eq:scoring_function}
  f(p) = \sum_{i=0}^{n} [logP_{AM}(l_i) + LMweight \cdot logP_{LM}(w(l_i)|w(l_0) \ldots
  w(l_{i-1})) - logP_{IP}]
\end{equation}
where:\\
$\bullet$ $logP_{AM}(l_i)$ is the acoustic model log-likelihood assigned to link $l_i$;\\
$\bullet$ $logP_{LM}(w(l_i)|w(l_0)\ldots w(l_{i-1}))$ is the language model
log-probability assigned to link $l_i$ given the previous links on the partial path $l_0 \ldots l_i$; \\
$\bullet$ $LMweight>0$ is a constant weight which multiplies the language model score 
  of a link; its theoretical justification is unclear but experiments
  show its usefulness;\\
$\bullet$ $logP_{IP}>0$ is the ``insertion penalty''; again, its theoretical
  justification is unclear but experiments show its usefulness.

\noindent To be able to apply the $A^*$ algorithm we need to find an appropriate 
stack entry scoring function $g_L(x)$ where $x$ is a partial path and
$L$ is the set of complete paths in the lattice. Going back to the
definition~(\ref{eq:g_overestimate})
of $g_L(\cdot)$ we need an overestimate $g(x.y)=f(x) + h(y|x) \geq
f(x.y)$ for all possible $y=l_k \ldots l_n$ complete continuations of
$x$ allowed by the lattice. We propose to use the heuristic:
\vspace{-0.3cm}
\begin{eqnarray}
  h(y|x) = \sum_{i=k}^{n} [logP_{AM}(l_i) + LMweight \cdot
  (logP_{NG}(l_i) + logP_{COMP}) - logP_{IP}] \nonumber\\
  + LMweight \cdot logP_{FINAL} \cdot \delta(k<n)  \label{eq:h_function}
\end{eqnarray}
A simple calculation shows that if $logP_{LM}(l_i)$ satisfies:
$logP_{NG}(l_i) + logP_{COMP} \geq logP_{LM}(l_i), \forall l_i$
then $g_L(x) = f(x) + max_{y \in C_L(x)} h(y|x)$ is a an appropriate
choice for the $A^*$ stack entry scoring function.
In practice one cannot maintain a potentially infinite stack. 
The $logP_{COMP}$ and $logP_{FINAL}$ parameters controlling the
quality of the overstimate in~(\ref{eq:h_function}) are adjusted
empirically. A more detailed description of this procedure is
precluded by the length limit on the article.

\section{Experiments}

\noindent As a first step we evaluated the perplexity performance of the 
SLM relative to that of a baseline deleted interpolation 3-gram model trained
under the same conditions: training data size 5Mwds (section 89 of
WSJ0), vocabulary size 65kwds, closed over test set. We have linearly
interpolated the SLM with the 3-gram model: 
$P(\cdot)=\lambda \cdot P_{3gram}(\cdot) + (1-\lambda) \cdot P_{SLM}(\cdot)$
showing a 16\% relative reduction in perplexity;
the interpolation weight was determined on a held-out set to be
$\lambda = 0.4$.  
\begin{table}[htbp]
  \begin{center}
    \begin{tabular}{lccc}
      \multicolumn{3}{l}{Trigram + SLM}\\\hline
      $\lambda$ &  0.0  &   0.4  &   1.0  \\\hline
      PPL     &    116  &   109  &   130  \\
      \multicolumn{3}{l}{Lattice Trigram + SLM}\\\hline
      WER     &   11.5  &   9.6  &  10.6  \\
    \end{tabular}\vspace{-0.4cm}
    \mycaption{Test Set Perplexity and Word Error Rate Results}
    \label{tab:results}
  \end{center}
\end{table}
A second batch of experiments evaluated the performance of
the SLM for trigram lattice decoding\footnote{The lattices were
  generated using a language model trained on 45Mwds and using a 5kwds 
  vocabulary closed over the test data.}. 
The results are presented in Table~\ref{tab:results}.
The SLM achieved an absolute improvement in WER of 1\% (10\% relative)
over the lattice 3-gram baseline; 
the improvement is statistically significant at
the 0.0008 level according to a sign test. 
As a by-product, the WER performance of the structured language model
on 10-best list rescoring was 9.9\%.

\section{Experiments: ERRATA}

\noindent We repeated the WSJ lattice rescoring experiments reported
in~\cite{chelba99:nldb} in a standard setup. We chose to work on the
DARPA'93 evaluation HUB1 test set --- 213 utterances, 3446 words. The
20kwds open vocabulary and baseline 3-gram model are the standard ones
provided by NIST.

As a first step we evaluated the perplexity performance of the SLM
relative to that of a deleted interpolation 3-gram model trained under
the same conditions: training data size 20Mwds (a subset of the
training data used for the baseline 3-gram model), standard HUB1 open
vocabulary of size 20kwds; both the training data and the vocabulary
were re-tokenized such that they conform to the Upenn Treebank
tokenization. We have linearly interpolated the SLM with the above
3-gram model:
$$P(\cdot)=\lambda \cdot P_{3gram}(\cdot) +(1-\lambda) \cdot
P_{SLM}(\cdot)$$
showing a 10\% relative reduction over the perplexity
of the 3-gram model. The results are presented in
Table~\ref{tab:ppl_results}. The SLM parameter reestimation
procedure\footnote{Due to the fact that the parameter reestimation
  procedure for the SLM is computationally expensive we ran only a single
  iteration} reduces the PPL by 5\% ( 2\% after interpolation with
the 3-gram model).  The main reduction in PPL comes however from the
interpolation with the 3-gram model showing that although overlapping,
the two models successfully complement each other. The interpolation
weight was determined on a held-out set to be $\lambda = 0.4$. Both
language models operate in the UPenn Treebank text tokenization.
\begin{table}[htbp]
  \begin{center}
    \begin{tabular}{lccc}
      \multicolumn{3}{l}{Trigram(20Mwds) + SLM}\\\hline
      $\lambda$                             &    0.0  &   0.4  &   1.0  \\\hline
      PPL, initial     SLM, iteration 0     &    152  &   136  &   148  \\
      PPL, reestimated SLM, iteration 1     &    144  &   133  &   148\\ 
    \end{tabular}
    \mycaption{Test Set Perplexity Results}
    \label{tab:ppl_results}
  \end{center}
\end{table}

A second batch of experiments evaluated the performance of the SLM for
3-gram\footnote{In the previous experiments reported on WSJ we have
  accidentally used bigram lattices} lattice decoding. The lattices
were generated using the standard baseline 3-gram language model
trained on 40Mwds and using the standard 20kwds open vocabulary. The
best achievable WER on these lattices was measured to be 3.3\%,
leaving a large margin for improvement over the 13.7\% baseline WER.

For the lattice rescoring experiments we have adjusted the operation of
the SLM such that it assigns probability to word sequences in the CSR
tokenization and thus the interpolation between the SLM and the
baseline 3-gram model becomes valid.  The results are presented in
Table~\ref{tab:wer_results}. The SLM achieved an absolute improvement
in WER of 0.7\% (5\% relative) over the baseline despite the fact that
it used half the amount of training data used by the baseline 3-gram
model. Training the SLM does not yield an improvement in WER when
interpolating with the 3-gram model, although it improves the
performance of the SLM by itself.
\begin{table}[htbp]
  \begin{center}
    \begin{tabular}{lccc}
      \multicolumn{3}{l}{Lattice Trigram(40Mwds) + SLM}\\\hline
      $\lambda$                             &    0.0  &   0.4  &   1.0  \\\hline
      WER, initial     SLM, iteration 0     &   14.4  &   13.0  &  13.7  \\
      WER, reestimated SLM, iteration 1     &   14.3  &   13.2  &  13.7  \\
    \end{tabular}
    \mycaption{Test Set Word Error Rate Results}
    \label{tab:wer_results}
  \end{center}
\end{table}

\section{Acknowledgements}

\noindent The authors would like to thank to Sanjeev Khudanpur for
his insightful  suggestions.
Also thanks to Bill Byrne for making available the WSJ lattices, Vaibhava Goel for
making available the N-best decoder, Adwait Ratnaparkhi for making
available his maximum entropy parser, and Vaibhava Goel, Harriet Nock
and Murat Saraclar for useful discussions about lattice
rescoring. Special thanks to Michael Riley and Murat Saraclar for help
in generating the WSJ lattices used in the revised experiments.

{\small
  \bibliographystyle{plain}
  \bibliography{camera_ready}
}
\end{document}